# Hate-Speech Detection in Roman-Urdu

**Moin Khan, Khurram Shahzad, Kamran Malik**

Punjab University College of Information Technology

University of the Punjab, Lahore, Pakistan

linktomoin@gmail.com, khurram@pucit.edu.pk, kamran.malik@pucit.edu.pk

# Abstract

Social media penetration is continuously increasing. That is, there are billions of users having access to some social media. Users express their thoughts, share their feeling, and news. While majority of the users use these medium positively, there are cases where users may leave in appropriate contents, such as hate-speech, which is widely acknowledged as an offense. Given that billions of posts are made every hour, it is not possible to manually detect these hate-speech. To address these issues, researchers have developed several techniques for a majority of European languages. However, little attention has been paid to Asian languages. In particular, to the best of our knowledge no study has been conducted for hate-speech detection for Roman Urdu, which is widely used in the sub-continent.

In this study, we have scrapped nearly 100,000 tweets generated from South Asia that can potentially have hate-speech content. Subsequently, we manually parsed the tweets to identify the Roman-Urdu tweets. After that, we employed an iterative approach to generate guidelines for hate-speech detection and used these guidelines to generate a hate-speech corpus for Roman-Urdu. The developed corpus is composed of over 5000 tweets, most of which are hate-speech. In particular, firstly categorized tweets into hostile and neutral. Secondly, we categorized the hostile tweets into offensive and hate-speech. The evaluation of the developed corpus is performed using several supervised learning techniques. The results showed that Logistic Regression is the most effective technique for distinguishing between neutral and hostile sentences.

# 1. Introduction

Urdu language has over 170 million speakers around the world [1]. It is the official language of Pakistan which is a country of 220 million inhabitants [2]. Apart from Pakistan, Urdu is spoken in many countries around the world, including India, UK, USA, Canada and the Middle East [3, 4, 5]. However, there are several challenges associated with Urdu writing [6]. For examples, the Urdu alphabets cannot be written using an English keyboard. Similarly, Urdu language has 40 alphabets [7], whereas, the English language keyboards are designed to handle 26 alphabets [8]. Therefore, it is not conveniently possible to map Urdu alphabets to an English keyboard. Due to these challenges, most of the Urdu speakers communicate in Roman Urdu, in which Urdu language written using English alphabets.

Growing number of smart devices, which includes e-readers, smart phones, desktop & laptop computers, and tablets. Furthermore, the increased penetration of social media has prompted a sharp increase in the number of Roman Urdu users [9]. Users of these devices continuously use these mediums to share their feelings, views, remarks and recommendations about products, services, politics, and other artifacts.

Due to the increased penetration of technology and enhanced social interaction, due to social media, people are now more prone to hate-speech attacks than ever before [10, 11, 12, 13, 14, 15]. For instance, a person who a different political affiliation may insult a politician or express hatred about them on the social media. Most of these opinions, comments and reviews are shared publicly and hence can be used easily to further provoke others to do the same.

Therefore, it is desired to identify such text, and removing them quickly to stop any further incitement [16]. Note, Hate-speech detection in social media texts is widely acknowledged as an important Natural Language Processing task [17, 18, 19], which has several crucial applications like sentiment analysis [20], investigating cyber bullying [21] and examining socio-political controversies [22]. Recognizing the importance of the issue, many countries have legislated to curb hate speech [23], and it is recognized as an offence [24].

## 1.1 Motivation

Today, growing number of people intends to express their feels on the social media, and with every passing day, people are getting more comfortable in expressing their views on social networking sites [25, 26]. As communication is revolutionized due to the availability of these social networking sites, it has increased the risk of promoting hate base activities and increasing propagation of hate-speech which is like to result in increased crime [27].

Many countries, such as Germany, France etc., are investing in developing an effective mechanism to counter such hate content right when it's published, so that it doesn't leave a negative impact on other users and they don't get provoked for any kind of revenge [28, 29]. Consider the recent example of infamous mosque attack in that left 51 people dead in New Zealand [30]. The gunman video streamed on Facebook the mass shooting, as it happened. Facebook reacted and removed that profane and highly disturbing video, however the videos is accessible with some effort [31]. Similar, could be the case with hateful text content.

In recent years, some research work has been done in developing methods to identify online hate speech in an automated way in the domain of social media [32, 33]. Many researchers from Natural Language Processing (NLP) and Machine Learning (ML) domain have shared their efforts [34, 35, 36, 37]. Our survey of hate-speech detection has revealed that almost all of the research has been done in English language and little or no work has been done on low-resource Asian languages, such as Roman-Urdu.

The focus of this research is to develop a benchmark corpus that can be used for evaluation of hate-speech detection techniques, as well as for training of machine learning techniques for the identification of hateful content. In particular, we have identified the following key challenges:

- *Lack of guidelines for other languages:* Although several studies have been conducted to identify hate-speech in English and other European languages, however, no generic and standardized guidelines are available that can be used for the manual identification of such content for other languages.

- *Absence of Roman-Urdu dataset:* No hate speech dataset currently exists for Roman-Urdu. To that end, we have developed the first ever dataset that contains comments and phrases that are commonly used in our daily life on social media networks.

- *Lack of benchmarks:* As no previous study has been conducted in this domain for Roman-Urdu, so it was hard to evaluate our results with any of the existing benchmark results for comparison purposes.

We have aimed to provide a platform which is smart enough to distinguish between a hostile and a neutral comment. In order to curb hate-speech, it's necessary to distinguish it from the offensive speech.

## 1.2 Research Objective

The basic purpose of this study is to develop the first-ever hate-speech detection corpus for Roman Urdu. The corpus will be useful in training machine learning technique to detect and thereafter omit such content. The choice of Roman Urdu stems from the fact that no such study has been conducted for Roman Urdu language. Given the background the aim of this study are as follows:
a) develop the first-ever hate-speech corpus for Roman Urdu, b) develop guidelines that can be used for a consistent annotation of hate-speech sentences, and c) evaluate the effectiveness of supervised learning techniques for identifying hate-speech in Roman Urdu text.

## 1.3 Significance of the Study

Identifying hate speech in the presence of increasing amounts of online content is a challenging task. Therefore, it is of higher significance to define hate-speech and what factors constitute hate-speech.

Another common issue that is faced most of the time in hate-speech detection is that it's often mixed with offensive speech. However, there's a fine line between hate and offensive speech but no standard guidelines exist that elaborates the difference between the two. In this study, we have put considerable effort to distinguish between the two types of speeches.

## 1.4 Overview of Research Methodology

The following steps have been performed step by step in this research:

I. In this step, we scrapped around 90 thousand tweets based on twitter handles, hot trends, geographical locations and famous hashtags.
II. Manual filtering was performed on that the scraped data in order to identify Roman Urdu tweets. In particular, imagery data was removed along with textual data having smileys and hashtags.

III. Guidelines were designed through an iterative process, so as to distinguish between neutral and hostile sentences. Hostile sentences were further classified as offensive and hateful-based using clearly defined guidelines. Note, separate guidelines were developed for both simple and complex sentences.
IV. Tagged data was then used in machine learning classifiers to train them on the given dataset, so that it can be expedited in future for automated testing of comments in Roman Urdu.
V. In this step performance of each classifier was analyzed and compared with each other in order to explicate the benchmark classifier for such data.

Figure 1.1 presented an overview of the complete methodology that we have used in this study:

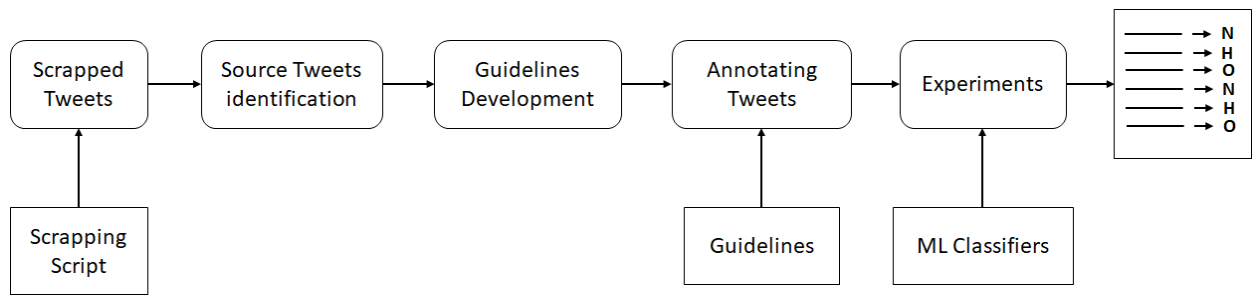

**Figure 1.1:** Hate Speech Detection Overview

## 1.5 Summary of Contributions

We have developed a hate-speech corpus for Roman-Urdu, using a systematic procedure. The procedure involves parsing the tweets and developing guidelines for annotation. The guidelines used in this process are aligned with the state of the art guidelines methodology given in the literature [38]. Below, we've presented the key benefits of the developed benchmark followed by a brief overview of each step of the process:

I. *Annotation of new unseen data:* The developed guidelines can serve as a baseline for the annotation of new data in Roman Urdu efficiently.
II. *Hate speech and the problem of Offensive language:* Offensive language is often mixed up with hateful content. We have comprehensively described the difference between the two. [39]

III. *Evaluation of proposed solution on standardized dataset:* Dataset has been developed from scratch in Roman Urdu language to evaluate and study the experimental results against the proposed metrics. Previously no such dataset existed to evaluate the performance for this domain.
IV. *Evaluation performance:* An experimental evaluation of the model on a Twitter dataset, demonstrating the top performance achieved on the hate speech detection task.

Figure 1.2 depicts the key steps that we have employed in this study. In essence, the proposed approach is composed of four major parts: Construction of corpus, application of proposed annotation guidelines, automated detection of hate speech and evaluation of experimental results.

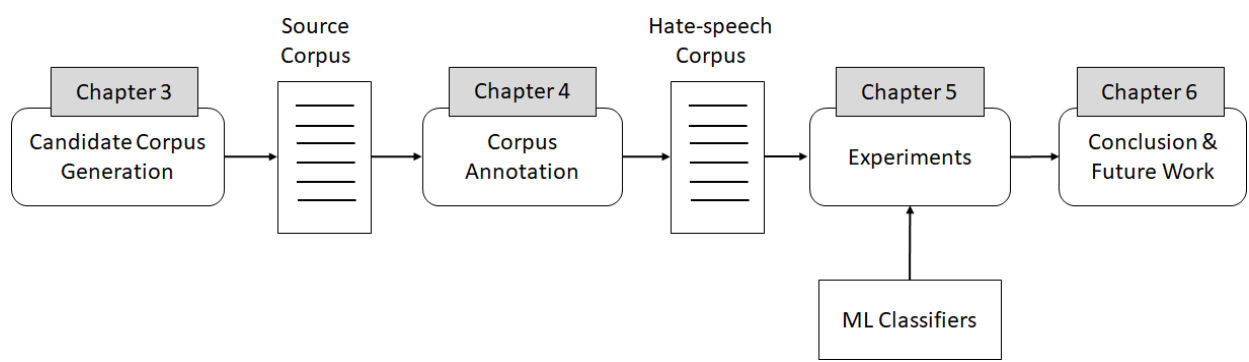

**Figure 1.2:** Development of Hate Speech Detection process

An overview of these actions is as follows:

- *Construction of Corpus:* Tweets were scraped from a famous social media site i.e. Twitter to develop the corpus in Roman Urdu. Twitter data scraping libraries were engaged for this purpose and scraping script was written in Python [40]. Tweets were filtered manually and only the most relevant to our cause were kept for the rest of the annotation procedure and experiments.

- *Application of Annotation Guidelines:* The proposed annotation guidelines were applied on the gathered and filtered dataset. This was done manually. In the first phase, normal comments were separated from the hostile ones. After that, another round was applied to the hostile comments to further classify them as Offensive or Hateful.

- *Automated detection of defined labels:* The tagged dataset was then supplied to the machine learning algorithms. The main purpose in employing this activity was to train a system efficient enough to detect hate speech in automated way by utilizing minimum human effort and give efficient results to support the cause. Machine learning classifiers were a big help in achieving this goal. The selected classifiers were trained on the annotated dataset using 10 fold cross-validations. Accuracies and F1 scores for each classifier was generated and saved for later analysis.

- *Evaluation of Experimental Results:* At the end, the performance of each classifier was compared to identify the most effective classifier.

# 2. Related Work

An established study [41] that investigated anti-black racism proposed that 86% of the time a tweet is categorized as racist is because of the reason that it contained offensive words. Keeping the high occurrence of offensive language and curse words in mind, this makes hate-speech detection a challenging task. Moreover, offensive words are often confusing as well, which makes the task even harder. For example, the word '*gay*' can be used both in the sense of contempt and in other contexts as well which are not related to hate-speech.

Malmasi and Zampieri in 2017 [42] worked on examining the methods which can be utilized to detect hate-speech in social media. The study presented a supervised classification system which relies on using n-grams and word skip-grams in their classifiers. Their dataset was comprising of English tweets having three annotation labels i.e. Hate, Offensive and OK (neither hate nor offensive). They achieved an accuracy of 78% with this experimental setting.

An attempt to detect hate-speech in Italian language was made by Del Vigna et al. in 2017 [43]. They built their dataset from the comments posted in Italian language on the public Facebook pages related to news, politicians, performers, groups and celebrities. Two different classification experiments were performed by them; the first one contained three different categories, which were *No Hate, Weak Hate and Strong Hate* [43]. For the second set of experiments, they combined Strong Hate and Weak Hate categories from the previous experimental settings and created one single class Hate; while the second class in this experiment was No Hate. They achieved accuracies of 65% and 73% respectively for the two experiments.

An important aspect in the hate-speech detection process is that it should not be mingled with offensive speech. It's not automatic criteria to tag a comment as abusive for just being offensive, as you can't hold someone accountable for just being offensive or sarcastic. People now-a-days are prone to use some terms which are highly offensive in their true meaning and historical background; but those are shared in a jolly, playful or sarcastic way. For example, the terms such as *hoe, bitch* and *nigga* are being used in everyday language of people or quoting rap lyrics or even for fun. Social media is a common source of sharing this kind of vocabulary, which makes this boundary condition very critical for any efficient hate speech detection process [44].

Badjatiya et al. in 2017 [45] employed deep learning in order to detect hate speech in tweets. Three neural network techniques were used in their study and the word embeddings in all the experimental processes were initialized with either word embeddings or GloVe embeddings. In their experiments, they used Long Short-Term Memory (LSTM), FastText and Convolutional Neural Network (CNN). The conducted experiments revealed that the performance of CNN was better than FastText and LSTM. Furthermore, when the word embeddings learned from these deep neural network models were grouped with gradient boosted decision trees, some significant improvements in the results were observed which significantly outperformed the existing methods.

An examination on the detection of hate speech using deep learning was conducted by Pitsilis et al. in 2018 [46]. An ensemble classifier was used and word frequency vectorization was exploited to extract the features which were later fed to the neural network based classifiers. Their results outperformed the existing methods and they concluded that no other model has classified short messages better, making their performance as a benchmark.

Nobata et al. in 2016 [47] strived to discern abusive language content using a supervised learning model which made use of several syntactic and linguistic attributes in the written text. Experiments were conducted at unigram and bigram level on the data collected from Amazon comments posted on Yahoo. A corpus of user comments tagged for abusive language content was also developed.

Waseem and Havoy in 2016 [48] proposed an unsupervised learning based solution, according to which a set of rules were proposed which a tweet should reveal if it has to be labelled as offensive. It was further showed in their study that different geographical distribution of users has minimum effect on the detection of offensive content.

An existing annotated dataset was taken by Waseem in 2016 [49] and a simple statistical analysis revealed the existence of notable connection between the likelihood of a user in sharing comments belonging to some offensive class i.e. *Sexism* or *Racism* and the tagged labels related with that class. The correlation coefficient value which was used to relate such likelihood of user was calculated to be 0.71 and 0.76 for racism and sexicm respectively on the mentioned dataset.

Arabic tweets were collected by Hamdy et al. (2017) [50] to create a newly created dataset having *Obscene*, *Offensive*, and *Clean* labels. List of obscene words was provided to tag each tweet as obscene if it contains a word from that list. Lists which were used in the study were SeedWords list, the LOR list (word bigrams only), the LOR list (word unigrams only), SeedWordslists + SeedWords lists and combined LOR (bigrams only) + combined LOR (unigrams only). They were able to achieve the highest F1 score of 0.60 on SW + LOR (unigrams).

Jha and Mamidi in 2017 [51] used the tweets dataset used by Waseem and Havoy (2016). They aimed to target classification problem of tweets but their scope was limited to *Sexism* only. They distributed the data into *Hostile*, *Benevolent* and *Other* categories. In contrast to the original dataset, they used Hostile instead of Sexism; and for Benevolent class, they collected their own tweet data. FastText and SVM classification algorithms were applied on the dataset.

A study done in German language [52] studied the obscene and vulgar categories by applying best-worst scaling. This technique is useful in the identification of emotional language and applying appropriate rating to it. They used the already existing lexicon having 3,300 items as seed lexicon and applied this technique to increase the lexicon size to 11,000 lexicon items.

The summary of results obtained by various researchers available in the literature to predict the Hate speech is presented below in Table 2.1.

| Author(s) Name | Language | Machine Learning Methods | Data Description | Performance |
|---|---|---|---|---|
| Malmasi and Zampieri (2017) | English | SVM with character n-grams, word n-grams and word skip-grams | 14,509 English tweets annotated by a minimum of three annotators | 78% accuracy |
| Del Vigna et al. (2017) | Italian | SVM & LSTM | Italian language comments posted on facebook | 65% and 73% for the two experiments |
| Davidson et al. (2017) | English | Logistic regression with L2 regularization | 24,802 tweets | 0.91 precision, 0.90 recall and 0.90 F1 score |

| Badjatiya et al. (2017) | English | FastText, CNN & LSTM | 16K annotated tweets | 0.93 precision, 0.93 recall and 0.93 F1 score |
|---|---|---|---|---|
| Hamdy et al. (2017) | Arabic | Unigram & Bigram with Log Odds Ratio (LOR) | 400K comments about politics, economy, society and science | 0.60 F1 score |
| Pitsilis et al. (2018) | English | RNN | 16,000 tweets | 0.89 F1 score |
| Nobata et al. (2016) | English | Character n-grams | Finance & News comments data | F1 score of 0.795 for Finance and 0.817 for News |
| Waseem and Havoy (2016) | English | Character n-grams | 16,914 tweets | 0.73 precision, 0.78 recall and 0.74 F1 score |
| Jha and Mamidi (2017) | English | SVM & FastText | 10,095 tweets | 0.87 precision, 0.87 recall and 0.87 F1 score |
| Eder et al. (2019) | German | Ridge regression with FastText embeddings | 3,300 lexical items | 11,000 lexical items |

**Table 2.1:** Previous work on Hate Speech Detection

# 3. Candidate Corpus Generation

In this work, we have developed the first-ever Roman Urdu corpus by using tweets. In particular, were scraped a large collection of tweets using twitter Python API that uses the advanced search option of Twitter Inc., and performed several preprocessing steps to generate a candidate corpus, which was subsequently annotated manually. Figure 3.1 shows the major steps that were performed for generating the candidate corpus, whereas, the details of the annotations are presented in Chapter 4. The details of each step of Figure 3.1 are as follows:

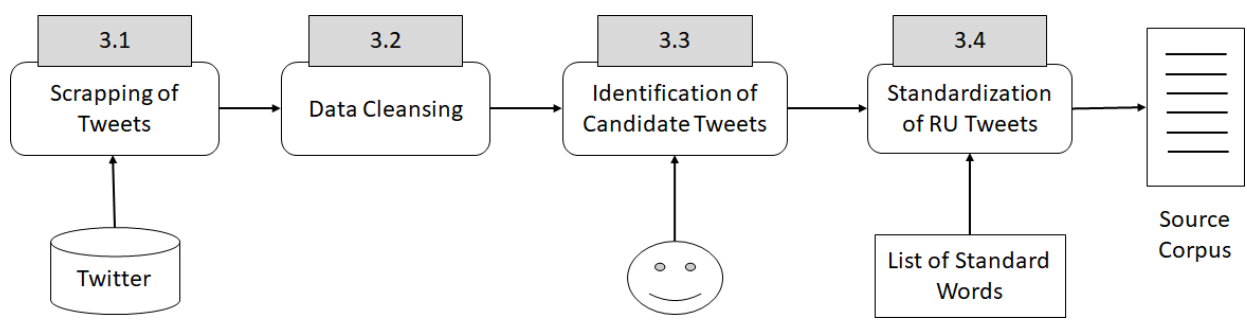

**Figure 3.1:** Candidate corpus generation procedure

## 3.1 Scrapping of Tweets

For our study, we decided to scrape the data from Twitter due to multiple reasons: a) Twitter is a large source of natural language text which is publically available, b) Twitter allows free scrapping of its tweets using a developer account and the APIs required to scrap it are available in wide range of languages, c) numerous existing studies have been done on twitter data, and d) Twitter has users all around the world, particularly its penetration in South Asia (where Urdu language is mostly spoken), hence provide a better representative of the offensive or hate-speech text written by the community.

Twitter's Python API was utilized for scrapping relevant tweets to generate a large collected of tweets that were subsequently used to generate the candidate corpus. In particular, the advanced search option of Twitter Inc., were used for filtering. That is, for scraping, we chose several keywords from politics, public protests, riots etc., that represents hot-trends, geographical locations and infamous hashtags from South Asia. For instance, #JamiaProtests, #ShameOnPolice, #BilawalBhonkoMut etc. The primary criteria for choice of these keywords were the possibility that these hashtags may likely to include offensive content, which is likely to include hate-speech. For example, #IndianHitlerModi hashtag is likely to have higher number of offensive or hate-speech content than #ChristmasDay or #EidMubarak. More importantly, we chose South Asia as a geo-location to increase the probability of including offensive content. Table 3.1 shows the types of hashtags and example hashtags of each type. It can be observed from the table that we have included hashtags from several genres, hence, making the source more diverse. We argue that that the finally developed hate-speech corpus will be useful for detecting hate-speech content from different genres.

| Type of hashtag | Example hashtag(s) |
| --- | --- |
| Politics | #PaisayDoAbbuLo, #AbbuBachao, #GhaddarMusharraf, #ModiMadness |
| Public protest | #IndiaRejectsCAA, #Emergency2019 |
| Riot | #JNUViolence |
| Religious | #JewSupremism |
| Racist | #PeopleOfColor, #RacistBJP |
| Scandal | #hareem_zaday, #BiggBoss13 |

**Table 3.1:** Example hashtags used for searching

Using the specifications of search given above, we scrapped over 90,000 tweets over a period of 3 months by choosing India and Pakistan as the source of tweet. Note that that tweets were scrapped in JSON format, which included unnecessary information such as images, timestamps, URL, smileys, hashtags, etc. An example set of source tweets is given below in Table 3.2. It can be observed from the comments in the table that the tweets include several contents that may not be useful in detecting hate-speech, thus requiring cleansing of the content. The details of the cleansing are discussed in the next step.

| Example Tweet | Comment |
| --- | --- |
| amir bhai aap bohut harami ho… @AamirLiaquat @siasatpk | It includes twitter username mentions |
| Trump eik safeed faam qoum parast hai jo dehshat gardi ko ubhaar raha hai :-( :-@ #DonaldTrump #TheClownMustGo #donaldtrumprealterrorist | It includes multiple emojis and hashtags |
| Plzzzzzzzz!!!!!! is dramy ko khatam na karein :(((( | It contains repetitive punctuation marks and misspelled words |

**Table 3.2:** Example tweets, as downloaded from Twitter

### 3.2   Data Cleansing

As discussed in the preceding step, the purpose of this step is to clean the text and extract the sentences that should be used for generating the candidate corpus. The details of the data cleansing steps are as follows:

***Removal of Imagery Content:*** A lot of imagery content, such as memes, trolls, GIFs, screenshots and photos, is shared on social media, and thus the scrapped tweet also included such content. As our study is targeting only the text data, so all those tweets having binaries of the images were removed from the dataset by using a Python script.

***Removal of URLs:*** All the links and URLs in the tweets were removed, as these do not contribute towards any kind of sentiment in the text. Furthermore, typically URLs do not contain meaningful sentences, therefore these contents were removed. Note, there were some tweets that only included a URL per se, these tweets were completely remove from our dataset.

***Removal of Usernames/Twitter handles:*** The scrapped tweets also contained user names, which disclosed the user who tweeted. Given that such as information is not useful for deciding whether the tweet is a hate speech or not therefore, this information was also omitted from the scrapped data. More specifically, we removed all such mentions, as a user's twitter handle or twitter username that have no significance in classifying sentences into hate speech or not.

***Removal of Emojis/Emoticons/Smileys:*** Emojis sometime help us to have a better understanding of user's emotions. But at times they can be misleading as well. For example, people often shares smile emoji ":-)" for sarcasm purpose. Therefore, in order to lessen the complexity and focusing more on the natural language text, we removed all the emojis from the dataset.

***Removal of Special Characters:*** Our screening revealed that several tweets casually use a bunch of punctuation marks or special characters in tweets to emphasize a feeling. However, these punctuation marks and special characters does not contribute to identification of offensive or hate speech content. Therefore, these contents were omitted from tweets using a Python script.

We preformed all the preprocessing steps discussed above to clean the scrapped tweets. Accordingly, the remaining content included natural language text of the tweets, which was further used in the next steps for generating the candidate corpus.

### 3.3 Identification of Candidate Tweets

A review of the content produced after cleansing revealed that a large majority of tweets were either neutral or they were not written in Roman Urdu, which is the focus of this research. Therefore, we thoroughly performed a semi-automated approach to identify a candidate corpus for manual annotation. Following sub-steps were carried out while performing this step.

- Firstly, we manually screened all the tweets to omit the tweets that were written in English or other European languages.
- Secondly, we also omitted the tweets that were replies to an existing tweet, retweets and the messages that were divided into multiple tweets, formally called tweet sequence. In this way, we made sure that each tweet can be treated independently.
- Finally, we identified the sentences that were written in Roman Urdu only and those were potentially offensive. Note, it does not mean that all the neutral sentences were omitted; we occasionally preferred non-neutral sentences over neutral sentences.

We preformed all the steps discussed above to handpick the most relevant tweets for our candidate corpus. Finally, we were remained with 5,000 tweets that were used in the rest of the study.

### 3.4 Standardization of Roman Urdu Tweets

It is widely acknowledged that there is not standard way of writing Roman Urdu [53]. Furthermore, there is no spellchecker for Roman Urdu, therefore, users tend to write their own way of spelling. For example, the word 'me' can be spelled in several ways which includes, main, mein, mai, mei, ma, and mn. A key reason for such a variation is that there is no one-to-one mapping exists as such between English letters and Urdu letters that are used for writing Roman Urdu text. Therefore, while developing a system that is capable of detecting hateful content in an automated way, all these spelling variations would be treated as different features by the training classifiers. Instead, all of these should be treated as single feature as most of the time they depict the same meaning. This could lead us to a more error-prone system, as testing classifiers may generate different results on these differently spelled words.

To address that problem, firstly we identified the vocabulary that is used in the collection of over 90,000 tweets. Secondly, we created a mapping between the commonly used vocabulary and its variations. Note that the meanings of the spelling were verified by taking into considering the whole tweet. Finally, the vocabulary was replaced with the commonly used spellings. Hence, a standardized dataset was achieved as an output of this process. We refer to the generated collection of tweets as candidate corpus. Table 3.3 illustrates the cleansing, identification of candidates and standardization processes.

| Raw Tweet | Standard Tweet | Potential Reason |
|---|---|---|
| Yeh tasveer bohut khubsurat hai | Yeh tasveer bohut khoobsurat hai | Standardized spelling of word "khoobsurat (beautiful)" has been applied |
| Mai karachi ja raha hun | Main karachi ja raha hun | Standardized spelling of word "main (me)" has been applied |
| Meri biwi ne mujhey apna kutta banaya huwa hai :-( :'( | meri biwi ne mujhey apna kutta banaya huwa hai | Removal of smileys |
| in hindu zalimon ko goli maar do :@ :@ :@ | in hindu zalimon ko goli maar do | Removal of smileys |
| in amreeki kutton ko afghanistan se nikal jana chahiye @realDonaldTrump | in amreeki kutton ko afghanistan se nikal jana chahiye | Removal of twitter username |
| khabardar jo tum ne mery samney bakwas ki!!!!!! | khabardar jo tum ne mery samney bakwas ki | Removal of special characters |
| I hate hindus | mujhey hindoun se nafrat hai | Translated to Roman Urdu from English |
| hum cpec per amreeki khudshaat se muttafiq nahi hein | hum CPEC per amreeki khudshaat se muttafiq nahi hein | Capitalizing an abbreviation (CPEC) to make it more recognizable |

**Table 3.3:** Examples of Standard Tweets

# 4. Corpus Annotation

In this section, we discuss the details of the steps that we have employed for generating the hate-speech corpus. To the best of our knowledge, no such corpus has been ever developed and thus not publicly available. Figure 4.1 shows the major steps that have employed. It can be observed from the figure that at first we developed an initial set of guidelines, and secondly, we employed an iterative approach to refine the guidelines. Finally, we applied the refined guidelines to generate our hate-speech corpus.

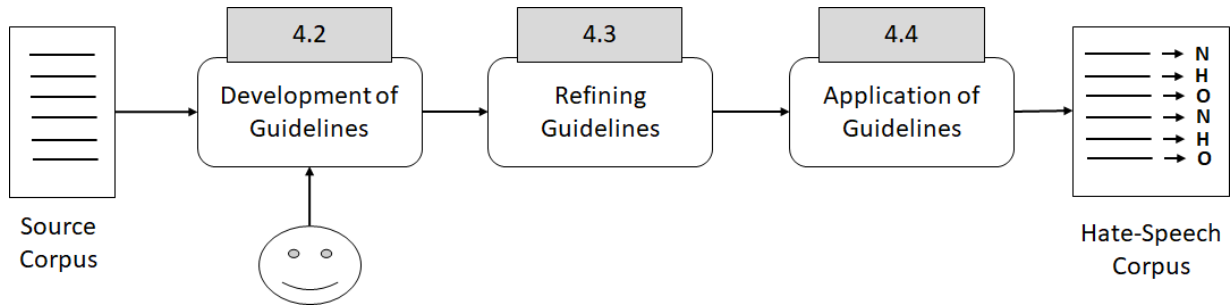

**Figure 4.1:** Generating hate-speech corpus

Prior to discussing the details of the developed corpus, we first discuss the details of the types of speech in Section 4.1. Subsequently, we present the details of the corpus development.

## 4.1 Types of Speeches

In this study, we have classified the sentences into three categories: *hate-speech, offensive, and neutral*. In general, hate-speech is a statement that may generate reaction or instigate revenge, given the chance, and it is targeted towards an individual or a community based on personal characteristics or the characteristics over which they have no control [54, 55, 56]. On the contrary, offense sentence reflects, anger, hatred, and may also break the relation, however, it does not instigate revenge, and does not target anyone based on personal characteristics.

More specifically, *hate-speech* is a hostile or sweeping statement that promotes hatred towards a community or a deprived group on the basis of gender, race, religion, ethnicity, origin, etc., over which the community has no control, and is likely to provoke an individual or a community for revenge [57, 58]. For instance, consider a comment "You are black and corrupt, as expected". The sentence is hate-speech as it reflects hatred towards a community due to a characteristic over he or she has no control. In contrast to hate speech, an *offensive* sentence is a hostile statement that bears the purpose of insulting or degrading an individual or a group of people without targeting any specific characteristic. For instance, a slight change in the hate-speech sentence example can make it an offensive, "You are corrupt, as expected" or "You all equally corrupt, as expected". Both expressions are disrespectful and hurtful towards an individual or a group, without being hate-speech. The third type of sentence, called *neutral*, is the sentence that nether contains hate-speech nor offensive language.

The further details of the three types of speeches are as follows:

### 4.1.1 Hate Speech

Following are some relevant definitions to define this category:

A hostile or sweeping sentence speech "that exhibits a clear desire to be cruel, promote hatred, incite harm, or attacks a group on the basis of characteristics such as religion, race, sex, national origin, ethnic origin, disability, sexual orientation, gender originality, or politics and may provoke them to take revenge" [56, 57, 58]. For example: "yeh musalman panah guzeen museebat saaz aur keery hein" (These muslim refugees are trouble makers and insects).

A comment that targets disadvantaged social groups "(groups of persons that experience a higher risk of poverty, social exclusion, discrimination and violence than the general population, including, but not limited to, ethnic minorities, migrants, people with disabilities, women, isolated elderly people and children) in a way that is possibly unsafe and harmful to them" [59]. For example: Jb b sarak per koi fakeer nazar aye, usey maar k bhaga do (Whenever you see any beggar on the road, beat him and make him run).

It is a speech that has a clear desire to promote hatred, or ignite harm by targeting individuals by indirectly referring to a group on the basis of their characteristics. For example: "Jo beghairat Mumtaz Qadri k khilaf bhonkay usey qatal ker do" (Any shameless person who barks against Mumtaz Qadri should be killed).

### 4.1.2 Offensive Speech

Following are some relevant definitions to define this category:

It is a speech that often conveys the purpose of insulting groups, and can include disrespectful, hurtful and abusive language. For example: Tamam sarkari hukkaam chor hain (All government servants are thieves).

It is a speech that is used to degrade another person and causes someone to feel angry, hurt, upset, insulting or rude [60]. For example: Miss Firdos Ashiq Awan ka makeup utaar do to ander se Mr. Firdos Khan nikal aye ga" (If you remove the makeup of Miss Firdos Ashiq Awan, she'll turn into Mr. Firdos Khan).

Unlike hate speech, it doesn't incite or inflict any direct harm to any individual person and doesn't target specifically on the basis of their characteristics. For example: "mein apne paisay jaisay chahon kharchon, tum kon ho kutia!!" (I'll spend my money how I want, who are you bitch!!)

### 4.1.2 Neutral

It is a statement that is being considered normal in everyday life. It is neither hateful nor offensive and depict a neutral intensity. For example:

a. Mein shiddat se agli episode ka wait kar raha hun (I'm impatiently waiting for the next episode).

b. Shaitan khudgarz aur paapi shaks k dil mein barh jata hai (The devil grows inside the heart of the selfish and wicked person).

### 4.2 Development of Guidelines

We have employed an iterative procedure to rigorously develop an exhaustive set of guidelines for determining whether a given expression is a hate-speech or not. A key benefit of developing these guidelines is that the use of these guidelines will lead to consistent and correct annotations.

To develop the guidelines, firstly, we randomly selected 10% sample from the candidate corpus and asked two annotators to annotate each tweets independently. Furthermore, the annotators were asked to develop an initial set of guidelines for annotation. The results of the annotations were compared, the difference were identified, and the guidelines were combined to generate a unified set of guidelines. Also, we computed inter-annotator agreement using Kappa statistic [61]. Typically, this is computed using *Cohen's inter-annotator agreement*.

Below, we present the initial set of guidelines to decide whether a given sentence is neutral, hate-speech, or offensive.

| *Neutral* |
|---|
| <ul><li>A sentence that leaves a pleasing feeling to every unbiased reader i.e. his/her affiliation, religion, or origin does not associate an emotional feeling.</li><li>A sentence that shares any information or knowledge or states a global fact, about himself, about any object, or the world in journal or a community which does not offend anyone.</li><li>Any quotation from a religious scripture or prophet, saints, gurus will be neutral.</li><li>A comment that is hostile or disrespectful but it does not specify a community or a group. Rather they use pronoun to refer to a subject.</li></ul> |
| *Offensive* |
| <ul><li>It doesn't incite or inflict any direct harm to any individual person or group and doesn't target specifically on the basis of their characteristics.</li><li>It is a speech that is used to degrade another person and causes someone to feel angry, hurt, upset, insulting or rude without being harmful in actual.</li><li>It is a speech that often conveys the purpose of insulting groups, and can include disrespectful, hurtful and abusive language.</li></ul> |
| *Hateful* |
| <ul><li>It is a speech that displays a clear desire to provoke harm, or to support hatred by targeting individuals or groups on the basis of their specific features.</li></ul> |

- It is a speech that victimizes disadvantaged social groups in a way that is possibly unsafe and harmful to them.

- It is a type of speech that shows a clear desire to be hurtful, to ignite harm, or to encourage hatred and attacks a group on the basis of attributes such as religion, race, sex, national origin, ethnic origin, disability, sexual orientation, gender originality, or politics and sometimes provokes them to take revenge.

**Table 4.1:** Initial set of guidelines

## 4.3 Refining Guidelines

The process was repeated a few times to refine the guidelines in such a way that higher inter-annotator-agreement can be achieved. As a result of the iterative process, we developed the taxonomy of the guidelines the concepts and developed guidelines for each element in the taxonomy. Figure 4.2 shows the taxonomy that we have developed.

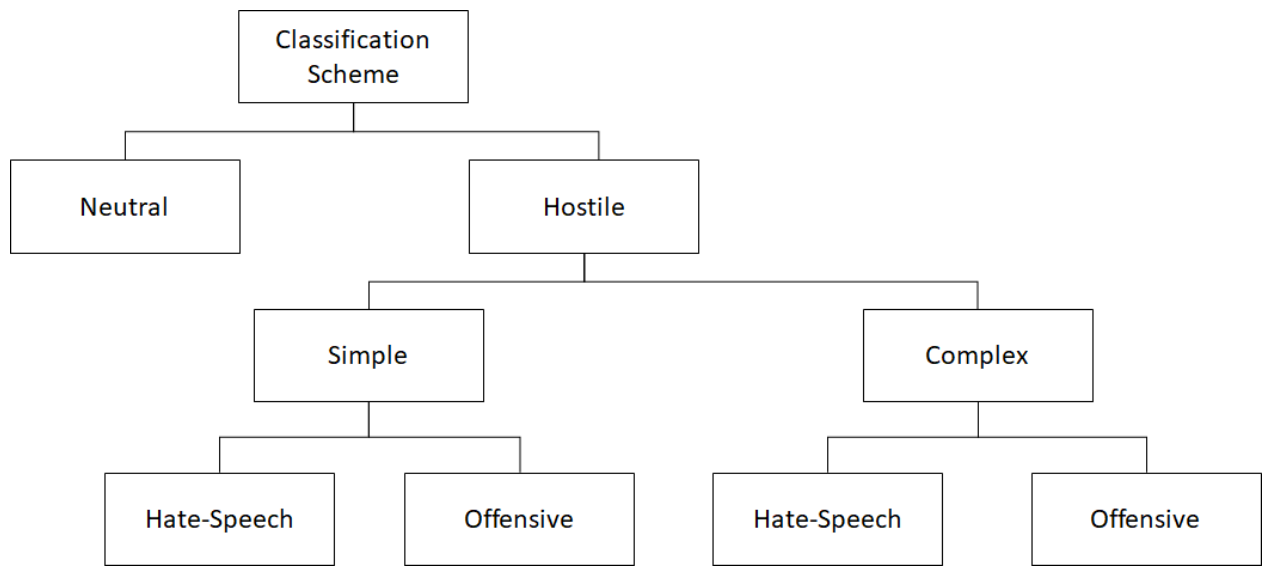

**Figure 4.2:** Taxonomy of the offensive sentences

For the verification of the guidelines we collected a total of 5,000 comments and randomly selected a 10% sample for evaluating how crisp are the guidelines. For generating sample, each comment was assigned an ID from 1 to 500, and we used python utility to generate 500 random numbers. Subsequently, two annotators independently used guidelines to classify the randomly selected 500 comments into hostile and neutral comments.

It was observed that the annotators had agreement on 96% of the observations. That is 393 were annotated as Hostile, 87 as neutral, 13 were declared as hostile by annotator 1 (Author) and neutral by annotator 2 (Neutral Annotator). Whereas, 7 were declared as neutral by A1 and hostile by A2. The details are as follows:

|       | A   | B  | Total |
|-------|-----|----|-------|
| A     | 393 | 7  | **400** |
| B     | 13  | 87 | **100** |
| Total | **406** | **94** | **500** |

**Table 4.2:** Kappa Calculations for Hostile/Neutral

No. of observed agreements: 480 (96.00% of the observations)
No. of agreements expected by chance: 343.6 (68.72% of the observations)

Kappa = 0.872
SE of Kappa = 0.028
95% confidence interval: From 0.817 to 0.927

| Inter annotator agreement Kappa score | 0.872 |
|---|---|
| Source Link | https://www.graphpad.com/quickcalcs/kappa2/ |
| No. of observed agreements | 96.00% of the observations |

**Table 4.3:** Kappa Score for Hostile/Neutral

Subsequently, for the second level of annotation, we took 25% of the previously selected sample data and asked two researchers to independently annotate the sentences, 393 tweets. That is 100 comments were selected in total for evaluating how crisp are the guidelines for hate and offensive speech categories. For generating sample, each comment was assigned an ID from 1 to 100, and we used python utility to generate 100 random numbers. Subsequently, two annotators independently used guidelines to classify the randomly selected 100 comments (out of those 393 comments) into hostile and neutral comments. After that, we computed inter-annotator agreement using Kappa statistic.

It was observed that the annotators had agreement on 91% of the observations. That is 15 were annotated as Hateful, 76 as Offensive, 2 were declared as hateful by annotator 1 (Author) and offensive by annotator 2 (Neutral Annotator). Whereas 7 were declared as offensive by A1 and hateful by A2. The details are as follows:

|       | A  | B  | Total |
|-------|----|----|-------|
| A     | 15 | 7  | 22    |
| B     | 2  | 76 | 78    |
| Total | 17 | 83 | 100   |

**Table 4.4:** Kappa Calculations for Hostile/Neutral

No. of observed agreements: 91 (91.00% of the observations)
No. of agreements expected by chance: 68.5 (68.48% of the observations)

Kappa = 0.714
SE of Kappa = 0.089
95% confidence interval: From 0.541 to 0.888

| Inter annotator agreement Kappa score | 0.714 |
|---|---|
| Source Link | https://www.graphpad.com/quickcalcs/kappa2/ |
| No. of observed agreements | 91.00% of the observations |

**Table 4.5:** Kappa Score for Hateful/Offensive

In order to validate the quality of annotation, we calculated the inter-annotator agreement (IAA) for the targeted speech annotation between the two annotation sets of 5000 Roman Urdu tweets using Cohen's Kappa coefficient. Kappa score is 0.872 and

0.714 respectively which indicates that the quality of the annotation and presented schema is productive.

Below, we present four sets of guidelines to decide whether a given sentence is neutral, hate-speech, or offensive. In particular, the first set of guidelines helps in distinguishing between neutral and hostile (offensive and hate-speech) comments. The second set of guidelines categorizes the hostile sentence into two types, simple and complex. The third set of guidelines can be used to declare simple sentence into offensive and hate-speech, whereas the fourth set of guidelines can be used to declare a complex sentence into offensive and hate-speech.

Note that the guidelines treat each sentence independent of its pre-context, as well as post-context. Therefore, all the sentences involving co-reference may not be considered as hostile to ensure the safe side of exclusion. The guidelines are as follows:-

| *Neutral* |
|---|
| **N1.** An expression that does not ignite anger or leave a pleasant feeling for an unbiased reader. <br> *Example: A*ap ke daant sitaron ki tarah hain (Your teeth are like the stars). <br><br> **N2.** That shares information, knowledge, or a fact, about a subject. <br> *Example:* Ethiopia k log ghareeb hain (Ethiopian people are poor). <br><br> **N3.** A quotation from a religious scripture, Prophets, Saints, or Gurus. <br> *Example*: Dunya mein aman qaim karo (Make the world peaceful). |
| *Hostile (Hateful/Offensive)* |
| **H1.** A hostile expression that promotes hostility, ignites anger, or incites harm to an individual or a small group. <br> *Example:* Tum eik ehsan faramosh kutia ho (You are an ungrateful bitch). <br><br> **H2.** A hostile sentence or information that portrays negativity or hatred about a community. <br> *Example:* Afghani dehshat gard hotay hain. (Afghans are terrorists). |
| **H3.** A misquote from religious scripture having hostile or disrespectful content.<br>*Example:* Islam ghair muslim logon ke sath imteyazi sulooq ka hukum deta hai (Islam commands to discriminate with non-Muslim people). |

**Table 4.6:** Guidelines to distinguish between Neutral and Hostile expressions

| *Simple* |
|---|
| **S1.** A sentence having a single clause. Where clause is a sentence portion containing both subject and verb.<br>*Example:* Us k baad us kutia ne shower lia (That bitch took shower after that).<br><br>**S2.** A direct speech expression.<br>*Example:* Mein usey apni kutia bulata hun (I call him my bitch).<br><br>**S3.** Does not contain any kind of Sarcasm. Where sarcasm is a taunt in which the author implicitly says something and everyone knows they mean the opposite.<br>*Example:* Mein usy apni kutia bulata hun (I call him my bitch). |
| *Complex* |
| **C1.** A sentence having more than one clauses.<br>*Example:* Wo kutia tung krti hai aur usey mernay ki zaroorat hai (That bitch is annoying and she needs to die).<br><br>**C2.** A sentence with indirect speech, quotation or reference (including retweets and posts sharing).<br>*Example:* Usne kaha k mjy us kutia se nafrat hai jo behas krti hai (He said that he hates a bitch that likes to argue).<br><br>**C3.** Contains sarcasm, ridicule or mockery.<br>*Example:* Tumhara chehra gadhey se milta hai (Your face resembles a donkey face). |

**Table 4.7:** Guidelines to distinguish between simple and complex expressions

| *Hateful* |
|---|
| **SH1.** That shows a clear desire to ignite harm, or to encourage hatred, by targeting individuals by indirectly referring to a group on the basis of their characteristics. <br> *Example:* Jo b Mumtaz Qadri k khilaf bhonkay usey qatal ker do (Kill anyone who barks against Mumtaz Qadri). <br><br> **SH2.** It is a speech that sweepingly victimizes disadvantaged social groups in a way that is possibly unsafe and harmful to them. <br> *Example:* In andhon ko dhakky mar k bahir nikalo (Push these blinds out). <br><br> **SH3.** That shows a clear desire to be hurtful, to ignite harm, or to encourage hatred and attacks a group on the basis of attributes such as religion, race, sex, national origin, ethnic origin, disability, sexual orientation, gender originality, or politics and sometimes provokes them to take revenge. <br> *Example:* Yeh musalman panah guzeen chor hain (These muslim refugees are thieves). |
| *Offensive* |
| **SO1.** It doesn't incite or inflict any direct harm to any individual person by indirectly referring to a group and doesn't target specifically on the basis of their characteristics. Example: qanoon ka samna karo jahil patwariyo (Face the law ignorant patwaris). <br><br> **SO2.** It is a speech that degrades a community and results in someone getting angry, hurt, upset, insulting or rude without being harmful in actual. <br> *Example:* Afghani namak harami krty hain (Afghans are thankless by nature). <br><br> **SO3.** It is a speech that often conveys the purpose of insulting groups, and can include disrespectful, hurtful and abusive language. <br> *Example:* Tamam sarkari hukmaran chor hain (All government officials are thieves). |

**Declaration:** The formulation of some of guidelines has been adapted from the existing work that is already published. These works have been cited in the table title given below.

**Table 4.8:** Guidelines to distinguish between hateful and offensive expressions in Simple comment [62, 63]

*MTO* = **Main target of opinion**

| *Hateful* |
|---|
| **CH1.** Explicit or implicit clue in the text suggesting that the speaker or author is in a state of aggression, hostile, antipathetic etc.<br>*Example:* Maar do, Taliban ko bhi aur unke hamiyon ko b (Kill Talibans and their supporters as well).<br><br>**CH2.** Explicit or implicit clue in the text suggesting that the speaker's attitude or judgment of the MTO is hateful i.e. speaker is frustrated, agitated, or very critical of the main entity.<br>*Example:* yahoodiyon ko qatal kerna sawab ka kaam hai!! (It is an act of virtue to kill jews!!)".<br><br>**CH3.** The MTO is considered predominantly hateful.<br>Example: Usama Bin Laden ne 9/11 attacks ki zimmadari qabool ki (Usama Bin Laden accepted the responsibility of 9/11 attacks). |
| *Offensive* |
| **CO1.** Explicit or implicit clue in the text suggesting that the speaker or author is in a state of anger, violent, irritated etc.<br>*Example:* aahh! Sb siyasatdan jhooty aur na-ahal hain (aahh! All the politicians are liars and incompetent).<br><br>**CO2.** Explicit or implicit clue in the text suggesting that the speaker's attitude or judgment of the MTO is offensive i.e. speaker is angry, disappointed, pessimistic, expressing sarcasm about, or mocking the main entity.<br>*Example:* Musharraf k masoom logon ko qatal krny per awam mein ghussa hai (People are angry on Musharraf for killing innocent people).<br><br>**CO3.** The MTO is considered predominantly offensive.<br>E*xample:* Jung k faislay ne lakhon logon ko be ghar kar diya (The war decision made many people homeless). |

**Declaration:** The formulation of some of guidelines has been adapted from the existing work that is already published. These works have been cited in the table title given below.

**Table 4.9:** Guidelines to distinguish between hateful, offensive and neutral expressions in Complex comment [38]

### 4.4 Application of Guidelines

Our developed annotation process is very crucial for correct and consistent annotation of an expression in a given sentence. Our application procedure is composed of the following guidelines:

A. Annotator should first determine whether a comment is *Neutral* or *Hostile* (Hateful/Offensive). Please refer to the guidelines in Table 4.6.

B. If the comment is Hostile, the annotator should identify whether that comment belongs to Simple or Complex comment category. Please refer to the guidelines in Table 4.7.

C. If the comment is simple, the annotator should identify whether the speech is about an individual, a small group, or the community at large. Subsequently, the annotator should follow the guidelines given in Table 4.8 for annotating the sentiment.

D. If the comment is complex, the annotation should use the guidelines given in Table 4.9. Please note that if both hateful and offensive parts are present in a complex comment, then it'll be declared as hate speech comment; hate speech being the superior class.

Apply the relevant guidelines from the table at each step. Means that if the speech is Hostile, belong to Simple category and about minorities, the guidelines about Disadvantaged Social Group becomes inapplicable. This has been depicted in the figure below (Figure 4.3).

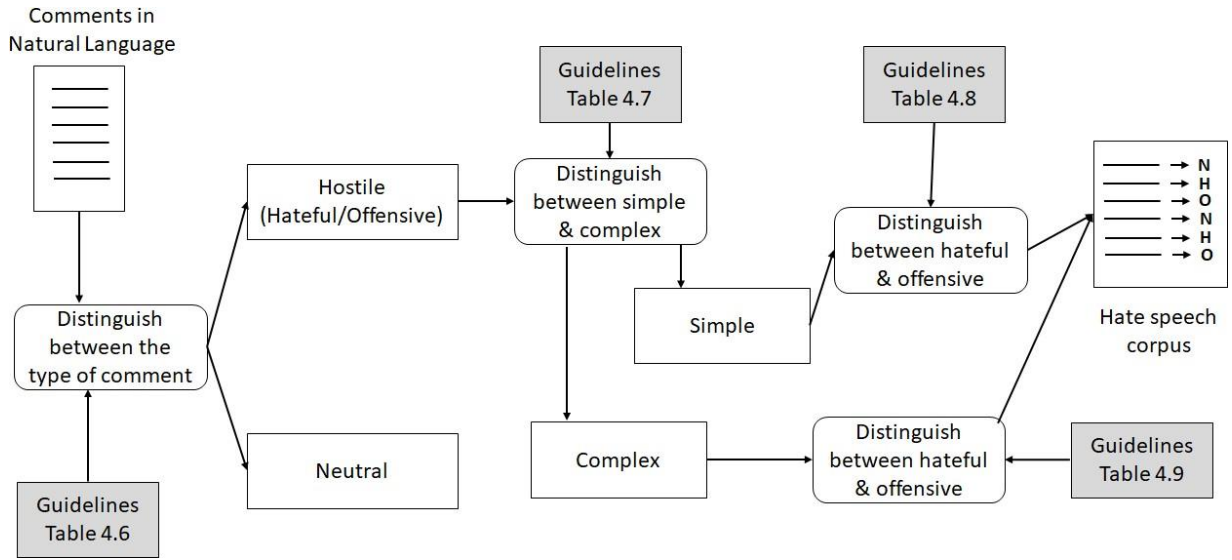

**Figure 4.3:** Application of Annotation Procedure

## 4.5 Specifications of the developed corpus

The annotated hate speech corpus (Neutral/Hostile dataset) consists of 5,000 Roman Urdu comments. Out of those 5,000 comments, 71% are *Hostile* and 29% belong to the *Neutral* category after applying the guidelines from Table 4.6. Those 71% Hostile comments are then further grouped together in a new dataset (Hateful/Offensive dataset) in which comments are classified into *Hateful* and *Offensive* categories with 75% and 25% ratio respectively after guidelines from Tables 4.7, 4.8 and 4.9 were applied on the Hostile comments accordingly. This has been summarized in Table 4.10 and Table 4.11.

| Comments in Neutral/Hostile Dataset: 5000 | | |
|---|---|---|
| **Type of Data** | **Comments Percentage** | **Comments Count** |
| Neutral | 29% | 1430 |
| Hostile | 71% | 3570 |

**Table 4.10:** Specification of Neutral/Hostile Dataset

| Comments in Hateful/Offensive Dataset: 3570 | | | | |
|---|---|---|---|---|
| **Type of Data** | **Comments Percentage** | **Comments Count** | **Hate-speech** | **Offensive** |
| Simple | 43% | 1547 | 308 | 1239 |
| Complex | 57% | 2023 | 525 | 1498 |

**Table 4.11:** Specification of Hateful/Offensive Dataset

# 5. Experiments

In this chapter we present the experimental settings that we have used to demonstrate the effectiveness of the developed corpus. In particular, the supervised learning techniques, feature selection techniques, which we have used for hate-speech detection in Roman Urdu. Subsequently, the experimental settings are discussed, the evaluation measures used for he evaluation are presented, and the experimental results are presented and analyzed.

## 5.1     Techniques and Features

To demonstrate the usefulness of the developed dataset for the hate-speech detection task, we have performed experiments using five supervised machine leaning techniques, including both classic and deep learning. These two types of techniques have been used in literature extensively [64, 65, 66]. There are some cases [67, 68] where classic techniques perform better than deep learning, for example while we have a small dataset for training usually classic techniques perform better and on larger dataset deep learning outperforms classical techniques [69, 70]. A brief description of each is given below.

### 5.1.1     Features Extraction

Text documents are a collection of words that are organized in specific order to convey a certain meaning. In order to use such files for learning and prediction, the text is extracted, parsed, pre-processed and converted into numbers, formally called feature vectors. These extracted features are given as input to machine learning techniques. The machine learning techniques takes these features as input and used them for learning and prediction. For instance, tf-idf features are further extracted based on word level, ngram level or character level features.

***Count Vectors as features:*** Count Vector is a de facto representation of features in matrix form. In the matrix every column is represents a term, whereas, every raw of the matrix represents a document. Hence, the value of each cell of the matrix represents the presence, absence or frequency of the term (representing the column) in the specific document. Every specific word in our dictionary will represent a feature (descriptive feature)**.** In count vectorizer method, feature vectors are created that learns the vocabulary of the dictionary and returns the Document-Term matrix (n_samples, n_features).

***Term Frequency vectors as features:*** Merely calculating the number of words in every document has limitations. It gives more importance in terms of weightage to longer documents than shorter documents. To avoid this, another method of term frequencies has been used i.e. count the number of times a word occurs in a document divided by the total number of words in each document.

***Tf-Idf vectors as features:*** In order to improve the results further, an improved version of term frequencies has been used i.e. Term Frequency Inverse Document Frequency (TF-IDF), in which higher weight is assigned to more common words that occur in all documents. The computed TF-IDF value shows the relative importance of the occurrence of a specific term in the document and consequently the entire dataset. It's calculated on the basis of two terms, Term Frequency (TF) score and Inverse Document Frequency (IDF). The first one calculates the normalized frequency score of a term. The latter one calculates the logarithm of the total number of documents in the dataset and divides it by the total number of documents where that particular term exists. The equations of TF and IDF are shown below:

$$\text{TF}(t) = \frac{\text{Number of times term t appears in a document}}{\text{Total number of terms in the document}} \qquad \text{Equation 5.1}$$

$$\text{IDF}(t) = \frac{\log_e(\text{Total number of documents})}{\text{Number of documents with term t in it}} \qquad \text{Equation 5.2}$$

This feature generation using TF-IDF vectors can be done at multiple levels e.g. word tokens, character tokens and n-gram tokens. Matrix created for word level TF-IDF shows the TF-IDF scores of every term present in all the documents of the dataset. Whereas the matrix created for N-gram Level TF-IDF shows the TF-IDF score of n-grams. N-grams are the combination of N terms together. Lastly, character Level TF-IDF matrix shows the TF-IDF scores of character level n-grams in the data.

***Word Embeddings:*** Word embeddings of a document are portrayed using a dense vector representation in which words of the data are mapped to vectors of real numbers. The position where a specific word will be placed within the vector is determined from the input text and is based on the surrounding words which are used in the input document. Input data itself can be used to train word embeddings but due to not having enormous data, word embeddings are generated using pre-trained word embedding of Word2Vec and used further for transfer learning**.** In order to use the word embeddings that are pre-trained, they are loaded and tokenizer object is created. Text data is then transformed to sequence if tokens and padded those tokens. These tokens are then mapped to their respective embeddings.

### 5.1.2    Supervised Learning Techniques

A plethora of supervised machine learning have been proposed in literature which have been used for various NLP tasks. A common utility function has been used that takes three inputs for the training of model and subsequently use them for future predictions. Firstly, name of the classifier; the feature vectors computed for the training data, thirdly, the assigned annotations of training data and finally the vector containing the features of validation data. We then trained the respective model using these inputs and subsequently accuracy and F1 scores are computed on validation data. The following classifiers are used to classify the data into neutral and hostile, hateful and offensive categories.

***Naive Bayes:*** Naive Bayes is a probabilistic classification method that is based on Bayes' Theorem [71]. A key feature of this classification approach is that relies on the assumption of creating conditional independence among predictors. It then assumes that the occurrence of a specific feature in a class is not related to the occurrence of any other feature.

***Linear Classifier:*** Logistic Regression has been implemented for linear classification. It is a statistical learning classification technique in which a logistic/sigmoid function is used to estimate probabilities of the categorical dependent variable value and checking it with one or more independent variables. The logistic function or sigmoid function generates a continuous input in the range of 0-1 which is helpful in probability.

***SVM Model:*** Support Vector Machine (SVM) are commonly used for both regression and classification tasks. This classification technique determines an optimal hyper-plane/line that separates the two output classes and uses this hyper-plane for prediction.

***Bagging Model:*** Bagging model that is employed here is Random Forest model. It is also commonly referred as Bootstrap Aggregating. The major advantage of using this ensemble and boosting classifier is its ability to reduce variance and avoid overfitting. Decision Trees are mostly used to generate output, which is combined to create bootstrap samples of the given input quantity and then the desired output is selected via voting.

***Boosting Model:*** Boosting models generate outputs by ensembling weak prediction models commonly decision trees in order to produce a prediction model. It builds its model like other ensembling models do in successive steps by allowing the optimization of an arbitrary differentiable loss function. Like other boosting models, this ensemble model also develops itself in step by step manner and generalizes the output by maximizing the arbitrary differentiable loss function. Boosting model that is implemented is Xtreme Gradient.

***Deep Neural Network Model:*** Deep Neural networks (DNN) is a type of neural network having greater complexity i.e. it involves multiple layers of neural network. Hidden layers are introduced in DNNs which contains neurons and these hidden layers are connected together for calculation of outputs and also for back propagations for error analysis. DNNs are capable of approximating linear as well as non-linear functions and hence can convert the input of any equation type into output. DNN model that is implemented in our work is CNN.

## 5.2     Experimental Setup

In this work, we have used the supervised learning techniques to evaluate their effectiveness of identifying hate-speech. For the experiments we divided the dataset into 90:10 ratios. That is, 90% of the randomly selected sentences were used for training of the selected classifiers, whereas the remaining 10% were used for testing. There were chances that the training set is biased, that is, there is a possibility that comments in training and testing might not be distributed normally for each class. To that end, we used 10-folds cross validation for each technique.

After preparing the hate-speech corpus, some preprocessing techniques were applied in order to improve the classification performance. This was done in order to obtain better results on the target dataset with selected classifiers. The major steps that were carried out during preprocessing were: case conversion, tokenization, parts of speech tagging and lemmatization.

For the detection of hostile speech in first part and offensive and hateful speech from hostile speech in the second part, the dataset includes 5000 and 3570 speech sentences, respectively. In case of hostile and neutral speech, the dataset is skewed towards hostile comments. Hostile class has 3570 speech sentences and neutral class has 1430 speech sentences. Labels of the classes are set to 'H' and 'N' in case of hostile and neutral class, respectively. In the latter case, i.e. offensive and hateful classification task, the speech sentences of offensive class are 2737 in number while speech sentences of hateful class are 833 in number. Labels of the classes are set to 'O' and 'H' in case of offensive and hateful class respectively. The dataset was divided into training and validation sets using the technique of k-fold validation. In order to run the training 10 times on different folds of the data, the value of k is set to 10.

Tensorflow and Keras are used to develop machine learning models and Convolutional Neural Networks along with other libraries: NLTK, Pandas, Numpy, Scikit-learn [72, 73, 74, 75]. An overview of our experimentation and evaluation phase is shown in the figure below (Figure 5.1).

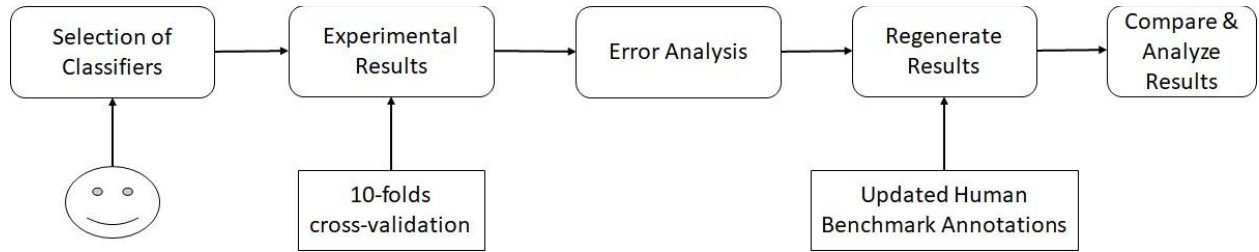

**Figure 5.1:** Automated Detection using ML Classifiers

Before consuming the dataset for feature extraction and classification, it requires some preprocessing in order to obtain better results. These steps are required in order to transfer text to machine-readable format, so that it can be used for further processing. In preprocessing, initially text is converted to lowercase and *tokenization* is performed, which is the process of breaking the given textual data into smaller chunks known as tokens. Typically, words, numbers, and punctuation marks are used to identify the token boundary. After tokenization is performed, stop words are removed from the data. *Stop words* are the commonly used words in the data. Since stop words do not carry any significant meaning, hence they are removed from the data. Another preprocessing step that is performed on the data is *lemmatization*. It is mainly performed on inflectional forms of the tokens and reduces them to a common base form. In contrast to the stemming process, inflectional forms are not straightforwardly dropped in lemmatization. That is, lexical knowledge bases are used to get the right base forms of words. In order to boost the results of classification, another important step performed is of parts of speech tagging. It aims to assign parts of speech (such as nouns, verbs, adjectives, and others) to each word of a given text based on its definition and its context. This preprocessed data is then fed to classification phase.

## 5.3 Evaluation Measures

We used three parameters Precision (P), Recall (R) and F1-score for the evaluation of the results. *Precision* is the number of correctly classified comments of a class divided by the numbers of comments classified as comments of that class. *Recall* is the count of named entities that are correctly classified divided by number of comments of that class in the test data. The mathematical formulas of Precision and Recall are given below.

$$\text{Precision}(P) = \frac{\text{Count of correctly classified comments of a class (true positive)}}{\text{Count of comments of that class in results (true positive+false positive)}} \qquad \text{Equation 5.3}$$

$$\text{Recall}(R) = \frac{\text{Count of correctly classified comments of a class (true positive)}}{\text{Count of comments of that class in test data (true positive+false negative)}} \qquad \text{Equation 5.4}$$

P does not represent the right evaluation, for example after an experiment, if we want to find P of hostile comments, suppose there were 100 hostile comments in test data, we found 10 hostile comments in results, all of 10 are correctly classified. The P of hostile comments becomes 1. However, the results are worst; 90 hostile comments were classified wrong. In this case R works, which will be 1/10. Likewise, R also fails. For example, in an experiment if we have a test data of 200 comments which includes 20 hostile comments. The results consist of 200 hostile comments out of 200 comments. The R of hostile comments will be 1, which is maximum; while the results are worst. In this case P works, which will be 1/10. To avoid problems like this harmonic mean of P and R is used which is called F1-score. *F1-score* is the optimal blend of precision and recall and we can combine the two metrics (P & R) in it. The mathematical formula of F score is given below in equation.

$$\text{F1-Score(F1)} = \frac{2*\text{Precision}*\text{Recall}}{\text{Precision} + \text{Recall}} \qquad \text{Equation 5.5}$$

## 5.4   Results

Results were calculated on the developed corpus using the above stated techniques. Table 5.1 shows the results of all the supervised learning techniques for their ability to distinguish between Neutral and Hostile comments. It can be observed from the table that the increase and decrease in results across folds is approximately comparable for each fold of Neutral/Hostile dataset. This represents that data was properly distributed for both classes in the training and testing folds. However, increase and decrease in individual results was different for each technique.

It can be observed from the table that Logistic Regression with Count Vectors emerged as the best technique. As we can see in the table, Logistic Regression with Count Vectors being a classic technique has performed better than Neural Network (CNN). This might be due to over-fitting problem in this deep learning technique.

We've further constructed the Table 5.2 to show the precision, recall and F1-score of each fold on the classification results of Logistic Regression with Count Vectors. F1-score across folds was highest at 1st fold i.e. 0.932, while lowest was at 9th fold i.e. 0.883. The variation among folds is almost streamlined and no major drop in the performance was observed at any fold. Average F1-score we've gotten is 0.906, which shows the effectiveness of our technique.

|  |  | Accuracy Score of 10-Folds | | | | | | | | | | Mean Score |
|---|---|---|---|---|---|---|---|---|---|---|---|---|
| Classifier | Feature | 1 | 2 | 3 | 4 | 5 | 6 | 7 | 8 | 9 | 10 | |
| NB | CLV | 0.86 | 0.80 | 0.78 | 0.85 | 0.83 | 0.82 | 0.77 | 0.79 | 0.75 | 0.80 | 0.80 |
| | NGV | 0.86 | 0.81 | 0.81 | 0.84 | 0.81 | 0.82 | 0.77 | 0.79 | 0.77 | 0.81 | 0.81 |
| | WLTF | 0.86 | 0.80 | 0.78 | 0.85 | 0.83 | 0.82 | 0.77 | 0.79 | 0.75 | 0.80 | 0.80 |
| | CV | 0.80 | 0.85 | 0.79 | 0.87 | 0.85 | 0.83 | 0.82 | 0.84 | 0.79 | 0.83 | 0.83 |
| LR | CLV | 0.88 | 0.81 | 0.80 | 0.85 | 0.84 | 0.82 | 0.77 | 0.80 | 0.76 | 0.80 | 0.81 |
| | NGV | 0.86 | 0.80 | 0.78 | 0.85 | 0.83 | 0.82 | 0.77 | 0.79 | 0.75 | 0.80 | 0.80 |
| | WLTF | 0.86 | 0.82 | 0.80 | 0.85 | 0.83 | 0.82 | 0.77 | 0.80 | 0.77 | 0.80 | 0.81 |
| | CV | 0.89 | 0.87 | 0.82 | 0.87 | 0.86 | 0.81 | 0.83 | 0.86 | 0.80 | 0.81 | 0.84 |
| RF | NGV | 0.86 | 0.81 | 0.81 | 0.84 | 0.81 | 0.82 | 0.77 | 0.79 | 0.77 | 0.81 | 0.81 |
| | WLTF | 0.86 | 0.82 | 0.78 | 0.85 | 0.84 | 0.81 | 0.80 | 0.80 | 0.79 | 0.82 | 0.81 |
| | CV | 0.85 | 0.80 | 0.80 | 0.81 | 0.84 | 0.81 | 0.77 | 0.80 | 0.78 | 0.83 | 0.81 |
| SVM | NGV | 0.86 | 0.80 | 0.78 | 0.85 | 0.84 | 0.82 | 0.77 | 0.79 | 0.75 | 0.80 | 0.80 |
| CNN | WE | 0.86 | 0.80 | 0.78 | 0.85 | 0.84 | 0.82 | 0.77 | 0.79 | 0.75 | 0.80 | 0.80 |

CLV = Character level vectors, NGV = N-gram vector, CV = Count vectors, WLTF = Word-level TF-IDF, WE = Word Embeddings, NB = Naïve Bayes, LR = Logistic Regression, RF= Random Forest, SVM = Support Vector Machine, and CNN = Convolutional Neural Network.

**Table 5.1:** Accuracies of Neutral/Hostile dataset

|  | Logistic Regression + Count Vectorizer | | | | | | | Avg. F1 score |
|---|---|---|---|---|---|---|---|---|
| Folds | TP | FP | FN | TN | P | R | F1 score | |
| 1 | 4160 | 500 | 110 | 230 | 0.893 | 0.974 | 0.932 | 0.906 |
| 2 | 3940 | 590 | 40 | 430 | 0.870 | 0.990 | 0.926 | |
| 3 | 3710 | 690 | 200 | 400 | 0.843 | 0.949 | 0.893 | |
| 4 | 4030 | 460 | 200 | 310 | 0.898 | 0.953 | 0.924 | |
| 5 | 4060 | 610 | 110 | 220 | 0.869 | 0.974 | 0.919 | |
| 6 | 3720 | 560 | 370 | 350 | 0.869 | 0.910 | 0.889 | |
| 7 | 3760 | 780 | 90 | 370 | 0.828 | 0.977 | 0.896 | |
| 8 | 3880 | 690 | 90 | 430 | 0.849 | 0.977 | 0.909 | |
| 9 | 3640 | 490 | 480 | 390 | 0.881 | 0.883 | 0.882 | |
| 10 | 3710 | 700 | 260 | 330 | 0.841 | 0.935 | 0.885 | |

TP = True Positive, FP = False Positive, FN = False Negative, TN = True Negative, P = Precision, and R = Recall.

**Table 5.2:** F1-Scores of Neutral/Hostile dataset

Table 5.3 shows the results of all the supervised learning techniques for their ability to differentiate between Offensive and Hate-speech comments. It can be observed from the table that the Hate/Offensive dataset when used with these techniques showed mean scores clustered in a close range, except CNN, which showed humongous difference in the results. This variation in the results of CNN is due to extreme over-fitting problem it might have faced during training, which caused the results during testing to deteriorate. Moreover, CNN was consuming the features generated by word embeddings and since the dataset was not so large, so the vectors trained on this dataset will presumably not carry much semantic information. So this technique was dropped from the analysis and we were left with classic techniques for our classification process. As mentioned, there isn't much difference in the mean score results across remaining techniques. However, we've observed from the table that the increase and decrease in results across folds was significant. Results of all techniques across folds were highest at 10th fold while lowest were at 5th fold.

This variation in results is due to improper distribution of comment categories in training and testing sets i.e. during random selection of training set, a major part of a class having specific features was ignored. While testing, the techniques will not be able to classify the comments from that class having those missed features, the results were decreased. On the other hand, the folds where a class having specific were distributed normally between training and testing, results were improved.

Logistic Regression with Count Vectors emerged as the best technique in terms of more accuracy. We've further constructed the Table 5.4 to show the precision, recall and F1-score of each fold on the classification results of Logistic Regression with Count Vectors. F1-score across folds was highest at 10th fold i.e. 0.812, while lowest was at 5th fold i.e. 0.692. Average F1-score we've gotten is 0.756. One can notice the significant increase in F1-Score of the lower folds. This depicts that data was properly distributed in the training and testing phases of those folds.

| Classifier | Feature | Accuracy Score of 10-Folds | | | | | | | | | | Mean Score |
|---|---|---|---|---|---|---|---|---|---|---|---|---|
| | | 1 | 2 | 3 | 4 | 5 | 6 | 7 | 8 | 9 | 10 | |
| NB | CLV | 0.69 | 0.66 | 0.80 | 0.78 | 0.82 | 0.77 | 0.80 | 0.87 | 0.83 | 0.85 | **0.79** |
| | NGV | 0.70 | 0.68 | 0.83 | 0.79 | 0.82 | 0.79 | 0.84 | 0.89 | 0.84 | 0.87 | **0.81** |
| | WLTF | 0.69 | 0.66 | 0.80 | 0.77 | 0.81 | 0.77 | 0.81 | 0.88 | 0.83 | 0.85 | **0.79** |
| | CV | 0.72 | 0.78 | 0.78 | 0.77 | 0.84 | 0.81 | 0.85 | 0.91 | 0.88 | 0.90 | **0.82** |
| LR | CLV | 0.71 | 0.69 | 0.82 | 0.81 | 0.84 | 0.80 | 0.81 | 0.87 | 0.85 | 0.86 | **0.81** |
| | NGV | 0.69 | 0.65 | 0.81 | 0.81 | 0.81 | 0.78 | 0.81 | 0.87 | 0.83 | 0.85 | **0.79** |
| | WLTF | 0.70 | 0.68 | 0.82 | 0.84 | 0.84 | 0.80 | 0.82 | 0.89 | 0.84 | 0.87 | **0.80** |
| | CV | 0.78 | 0.78 | 0.84 | 0.83 | 0.83 | 0.84 | 0.84 | 0.90 | 0.89 | 0.90 | **0.84** |

| | | | | | | | | | | | |
|---|---|---|---|---|---|---|---|---|---|---|---|
| **RF** | NGV | 0.71 | 0.68 | 0.83 | 0.77 | 0.86 | 0.78 | 0.83 | 0.88 | 0.83 | 0.88 | **0.80** |
| | WLTF | 0.73 | 0.71 | 0.82 | 0.82 | 0.84 | 0.81 | 0.85 | 0.89 | 0.88 | 0.86 | **0.82** |
| | CV | 0.75 | 0.72 | 0.82 | 0.82 | 0.83 | 0.84 | 0.84 | 0.90 | 0.85 | 0.86 | **0.82** |
| **SVM** | NGV | 0.69 | 0.66 | 0.80 | 0.78 | 0.81 | 0.77 | 0.80 | 0.87 | 0.83 | 0.85 | **0.79** |
| **CNN** | WE | 0.31 | 0.34 | 0.20 | 0.22 | 0.19 | 0.23 | 0.20 | 0.13 | 0.17 | 0.15 | **0.21** |

CLV = Character level vectors, NGV = N-gram vector, CV = Count vectors, WLTF = Word-level TF-IDF, WE = Word Embeddings, NB = Naïve Bayes, LR = Logistic Regression, RF= Random Forest, SVM = Support Vector Machine, and CNN = Convolutional Neural Network.

**Table 5.3:** Accuracies of Hate/Offensive dataset

| | **Logistic Regression + Count Vectorizer** | | | | | | | Avg. F1 score |
|---|---|---|---|---|---|---|---|---|
| **Folds** | **TP** | **FP** | **FN** | **TN** | **P** | **R** | **F1 score** | |
| 1 | 880 | 60 | 700 | 2150 | 0.936 | 0.557 | 0.698 | |
| 2 | 980 | 120 | 620 | 2070 | 0.891 | 0.613 | 0.726 | |
| 3 | 690 | 120 | 370 | 2610 | 0.852 | 0.651 | 0.738 | |
| 4 | 750 | 170 | 390 | 2480 | 0.815 | 0.658 | 0.728 | |
| 5 | 630 | 170 | 390 | 2600 | 0.788 | 0.618 | 0.692 | **0.756** |
| 6 | 890 | 200 | 290 | 2410 | 0.817 | 0.754 | 0.784 | |
| 7 | 580 | 70 | 200 | 2940 | 0.892 | 0.744 | 0.811 | |
| 8 | 810 | 240 | 250 | 2490 | 0.771 | 0.764 | 0.768 | |
| 9 | 690 | 80 | 250 | 1770 | 0.896 | 0.734 | 0.807 | |
| 10 | 622 | 60 | 228 | 1880 | 0.912 | 0.732 | 0.812 | |

TP = True Positive, FP = False Positive, FN = False Negative, TN = True Negative, P = Precision, and R = Recall.

**Table 5.4:** F1-Scores of Hate/Offensive dataset

We've examined from the above tables that the average of Logistic Regression + Count Vectorizer across ten folds is 0.906 and 0.756 for Neutral /Hostile and Hate/Offensive datasets respectively with each of ten folds. F1-scores of each fold for both datasets have been displayed in the Figure 5.2 below.

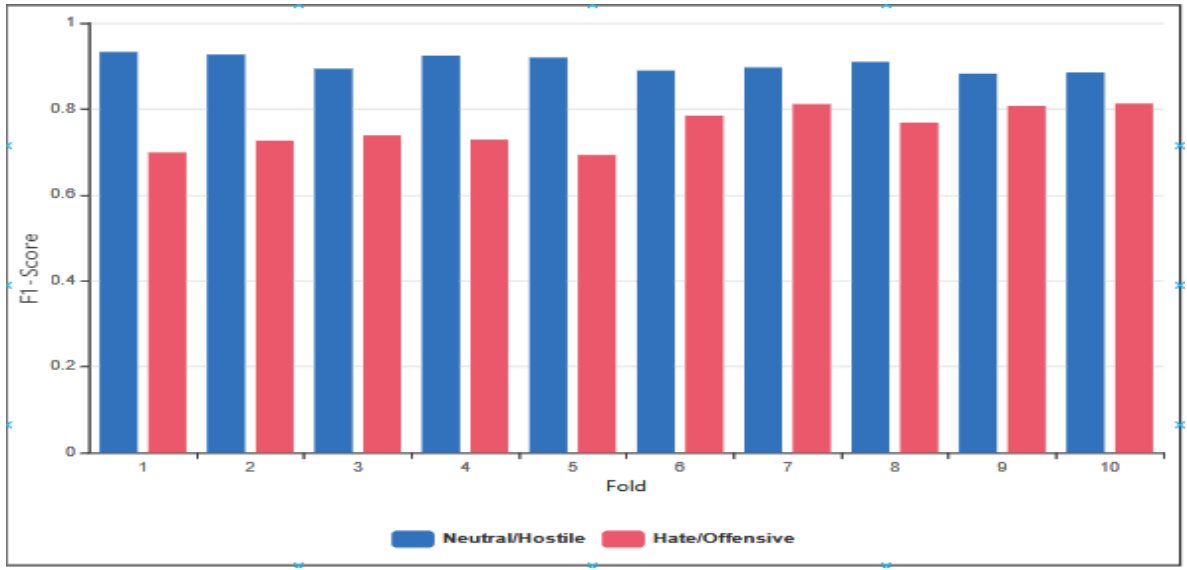

**Figure 5.2:** F1-scores of Neutral /Hostile and Hate/Offensive datasets

## 5.5 Error Analysis

In order to analyze the underlying reasons of misclassification during experimentation, we filtered all the examples of the sentences that were incorrectly classified. A potential reason was attempted to sought for the misclassification during this activity. Table 5.5 shows the example sentences, actual label, predicted label and identity of the reasons for classification, whereas, Table 5.6 shows the details of the reasons.

| Sentence | Actual Label | Predicted Label | Reason |
|---|---|---|---|
| eik nasli **kutta** 100 **badnasal** yaron se behter hota hai | Neutral | Hostile | A |
| Pakistan ka **president** hamesha **janwaron ki dunya** se select kiya jata hai | Hostile | Neutral | B, C |
| eik achi larki ka dost hona eik **randi** ke dost honay se bohut behter hai | Neutral | Hostile | A |
| teray khoon mein **beghairti ki miqdaar** bohut zayada hai | Hostile | Neutral | G |
| JUI **islam card** istemaal ker rahi hai | Hostile | Neutral | E |
| uski **khabasat** ka level check karo | Hostile | Neutral | G |
| showbaz sharif har toofani barish ke baad long boot pehan ker aisey bahir nikal aata tha, jesay barsaat mein mendak nikaltey hein | Hostile | Neutral | C |
| agar insaan ki pehchaan uske doston se hoti hai tou main is duniya ka sab se bara beghairat insaan hun | Hostile | Neutral | C |
| pehly giraftaariyan pathar maarney pe hoti theen, ab sirf pathar maarney ke shabay mein dhar liya jata hai | Hostile | Neutral | D |
| **chughalkhor** eik lamhey mein itna **fasaad** barpa ker deta hai jitna jadugar eik mahiney mein nahi ker sakta | Neutral | Hostile | A |
| maryam ka papa papa nai papi hai | Hostile | Neutral | C, E |
| Yar dil kr raha hai sb police stations ko bomb sy uraa dy banda | Hostile | Neutral | D |
| Khatoon driver ne **Allaho Akbar** ka naara lagaya aur 40 logon per gaari dhora di | Hostile | Neutral | J |
| hakumat aur pak fouj se guzarish hai k **TTP** aur **ISIS** k logon ko din ki roshni mein pakrein aur andhera honay se pehlay latka dein | Hostile | Neutral | F |
| **Allah** aisy jahil logon ko kabi na bakhshy | Hostile | Neutral | I |
| Ghareebon ko to khatam kr dena chahiye | Hostile | Neutral | B |
| **Trump** ne Venezuela per nai muashi pabandiyan laga di hein | Neutral | Hostile | H |
| bharat se rozana bakriyon aur kutton se bhi jinsi **zeyadati** ki khabren aati hein | Hostile | Neutral | J |
| Police walo tuaddi **pein di sirri** | Hostile | Neutral | G |
| seyasi maqsad hasil kernay ke liye **tashaddud** kerna **dehshatgardi** hai | Neutral | Hostile | A |

**Table 5.5:** Examples of Misclassified Examples

| Label of Reason | Reason |
|---|---|
| A | Use of Hate word(s) |
| B | No Hate word(s) |
| C | Sarcasm/Taunt |
| D | Offensive Act (without much hate wording) |
| E | Blame/Threat (without much hate wording) |
| F | Abbreviation of some offensive/hateful term is used |
| G | Hate word rarely occurred in the dataset |
| H | Word(s) is predominantly hateful in the dataset |
| I | Word(s) is predominantly neutral in the dataset |
| J | Spelling variations in Roman Urdu |
| **Note:** Partially Hostile and partially Neutral or partially Hateful and partially Offensive will be tagged as Hostile and Hateful respectively | |

**Table 5.6:** Potential Identified Reasons

# 6. Conclusion & Future Work

In this study, we have taken the first-ever step towards hate-speech detection for Roman Urdu. While numerous such attempts have been made to several Asian, as well as European languages, as far as we are aware, no attempt has been made for Roman Urdu.

In particular, we have developed the first-ever hate-speech corpus for Roman Urdu and evaluated the effectiveness of several supervised learning techniques. For the development of the corpus we scrapped nearly 100,000 tweets using several hashtags having South Asia as origin. Subsequently, the tweets were cleaned to omit web links, fragmented tweets, replies to existing tweets, and images. The cleaned corpus was manually reviewed to identify the tweets that were written in Roman Urdu, we refer to it as candidate corpus.

To annotate the candidate corpus, we employed an iterative approach to develop concreted sets of guidelines. The first set of guidelines can be used to distinguish between offensive and neutral tweets; the second set of guidelines separates simple sentences from complex sentences. The third set of guidelines were developed for simple sentences to differentiate between offensive and hate-speech sentences. The final set of guidelines were developed for complex sentences to differentiate between offensive and hate-speech sentences. The guidelines were used to annotate the candidate corpus in order to generate the hate-speech corpus.

In this study, we also employed multiple supervised learning techniques to evaluate their ability to identify offensive sentence, categorize between simple and complex sentences, and also categorize offensive and hate-speech sentences. We also developed the guidelines for the separation of hate-speech and offensive speech. The results of the experiments revealed the following; a) Logistic Regression is the most effective technique for distinguishing between neutral and hostile sentences and Count Vectors are the most effective features, b) Logistic Regression with Count Vectors is also the most effective technique for distinguishing between hateful and offensive sentences, c) Character n-grams and Word n-grams didn't perform quite well on the developed corpus, majorly due to the spelling variation issue in Roman Urdu, d) sarcasms were really hard to detect because at times no hateful or offensive word was used in the comment, which makes the polarity of the comment as neutral while classifying it.

Following are the potential areas of future research: a) guidelines can be made more generic in order to cater the needs for other languages as well, b) the dataset needs to be extended for better training and learning, c) further detailed analysis can be done on morphological and syntactic rules; and using character n-grams, unigrams and bigrams to deal with the rich morphology of Urdu, and d) handling of smileys in the comments could help us understand the sentiment in a better way.